\documentclass[times,twocolumn,final,authoryear]{elsarticle}

\usepackage{prletters}
\usepackage{framed,multirow}

\usepackage{amssymb}
\usepackage{latexsym}

\usepackage{epstopdf}
\usepackage[caption=false]{subfig}
\usepackage{multirow}
 \usepackage{url}
\usepackage{amsmath}
\usepackage{graphicx}
\usepackage{array}
\usepackage[linesnumbered,ruled]{algorithm2e}

\newcolumntype{M}[1]{>{\centering\arraybackslash}m{#1}}
\DeclareMathOperator*{\argmin } {arg\, min}

\usepackage{url}
\usepackage{xcolor}
\definecolor{newcolor}{rgb}{.8,.349,.1}

\journal{Pattern Recognition Letters}

\begin{document}


\ifpreprint
  \setcounter{page}{1}
\else
  \setcounter{page}{1}
\fi

\begin{frontmatter}

\title{Are you eligible? Predicting adulthood from face images via class specific mean autoencoder}

\author[]{Maneet \snm{Singh}} 
\author[]{Shruti \snm{Nagpal}}
\author[]{Mayank \snm{Vatsa}\corref{cor1}}
\cortext[cor1]{Corresponding author: 
  Tel.: +91-9654653404;  
  fax: +91-11-26907410;}
\ead{mayank@iiitd.ac.in}
\author[]{Richa \snm{Singh}}

\address{IIIT-Delhi, New Delhi, 110020, India}

\received{15 March 2017}

\begin{abstract}
Predicting if a person is an \textit{adult} or a \textit{minor} has several applications such as inspecting underage driving, preventing purchase of alcohol and tobacco by minors, and granting restricted access. The challenging nature of this problem arises due to the complex and unique physiological changes that are observed with age progression. This paper presents a novel deep learning based formulation, termed as Class Specific Mean Autoencoder, to learn the intra-class similarity and extract \textit{class-specific} features. We propose that the feature of a particular class if brought similar/closer to the \textit{mean} feature of that class can help in learning class-specific representations. The proposed formulation is applied for the task of \textit{adulthood} classification which predicts whether the given face image is of an adult or not. Experiments are performed on two large databases and the results show that the proposed algorithm yields higher classification accuracy compared to existing algorithms and a Commercial-Off-The-Shelf system.
\end{abstract}


\end{frontmatter}

\section{Introduction}
\label{sec:intro}
Human aging is a complex process and brings with it behavioral and physiological changes. One associates maturity and mental growth with the behavioral changes that occur with time. Age is also often used as a means of access control, physically as well as virtually, to keep younger minds away from activities and content they are not deemed ready for. A threshold age, known as the \textit{age of majority}, is defined by most states to universalize the concept of an individual being physically and mentally ready to assume control for their actions and decisions. However, there are different age limits prescribed by individual state and federal governments for different activities. For instance, in the USA, the legal age for smoking is 18 years while age limit for voting and drinking is 21 years.

\begin{figure}
\centering
\subfloat[Below the age of majority]{\includegraphics[height= 1.2in]{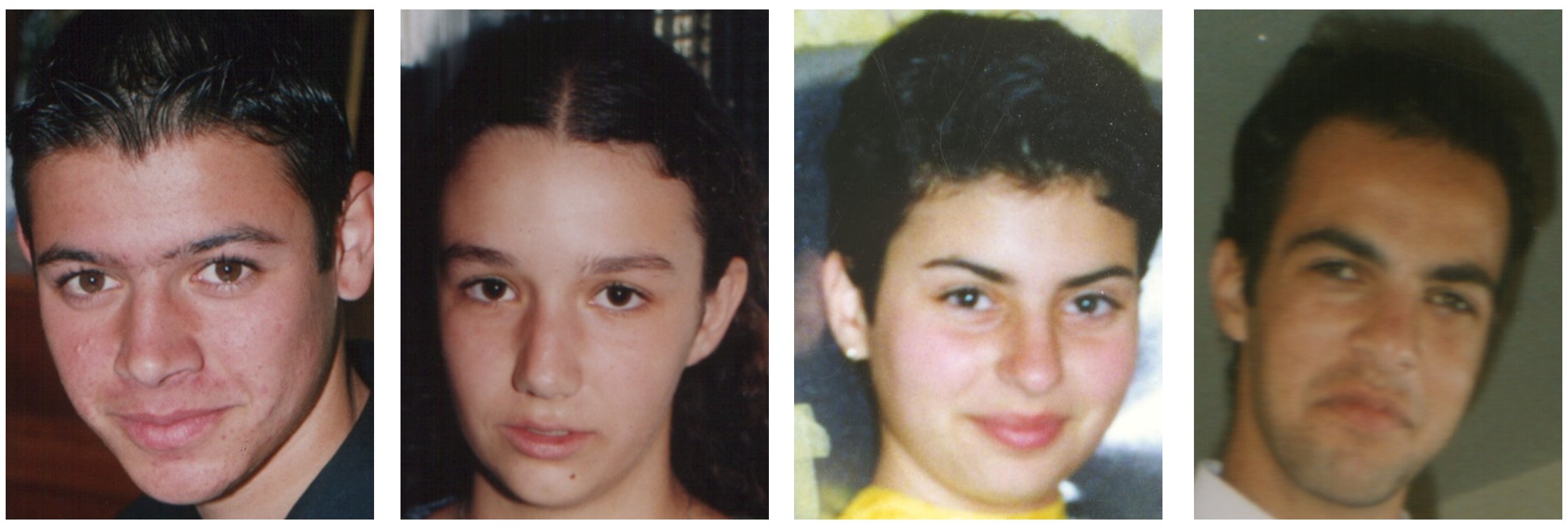}}  \\
\subfloat[Above the age of majority]{\includegraphics[height= 1.2in]{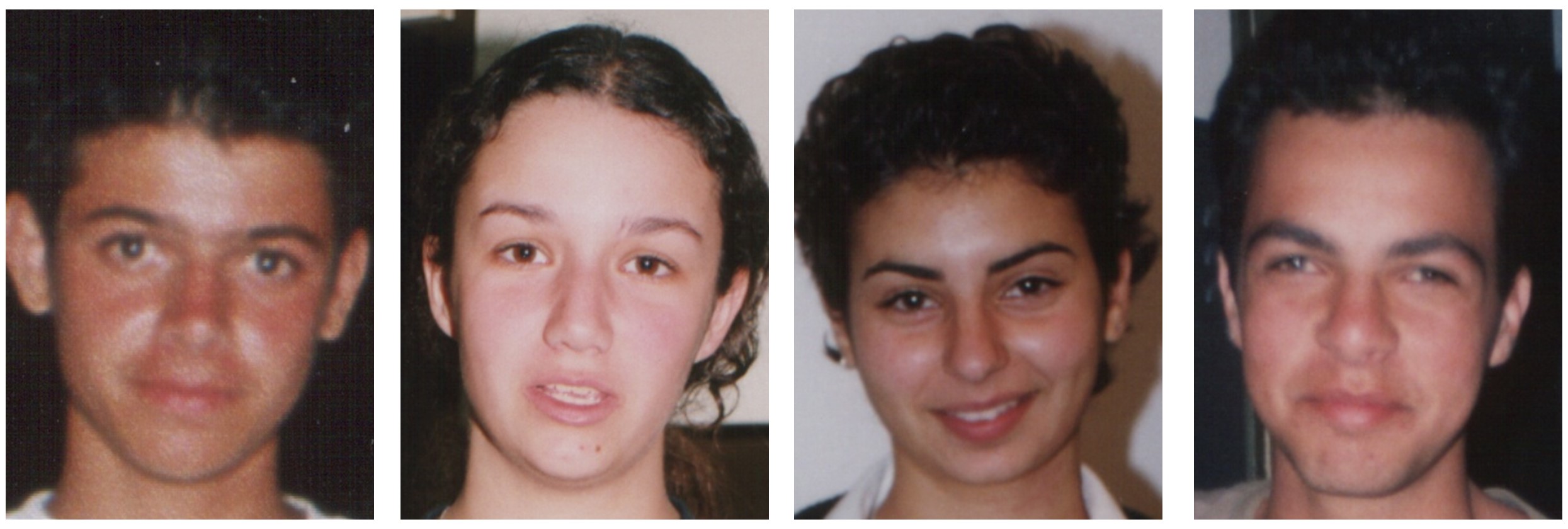}}  
\caption{Sample images from the FG-NET Aging dataset \citep{fgnet}. (a) shows images of individuals below the age of majority, and (b) shows sample individuals at the age of majority. These examples illustrate the challenging nature of \textit{adulthood classification}. }
\label{fig:intro}
\end{figure}

The physiological and behavioral effects of aging vary for every individual and are a function of several parameters such as health, living style, environmental conditions, and gender. Therefore, it is challenging to accurately estimate the age of a person. As can be seen from Fig. \ref{fig:intro}, it becomes difficult to predict the age of individuals just immediately below, or above the age of majority (e.g. 18 years). Among the currently available non-intrusive biometrics, the face \textit{changes} significantly with age and is therefore a preferred modality for estimating the age of a person. As of now, at different checkpoints, an officer or a designated person in-charge estimates the age of a person by visually observing an individual. In cases where visual inspection becomes difficult, she/he asks for an identification (ID) card as the proof of age. However, research has shown that both these measures used for age estimation and verification are prone to errors \citep{fake, juvenile}. With easy availability of tampered or fake ID cards, several youngsters use these cards to mis-represent their identity. Recently, in a survey at Harvard University \citep{underage}, it was found that $18\%$ underage students obtained alcohol using fake ID cards. Inspired by these observations and motivated by the increasing use of technology and automation in our day-to-day life, this research aims to automate the process of classifying an individual as an adult or not. For a given face image, the proposed algorithm aims to predict if the person has attained the age of majority or not. Such a system could be deployed at multiple places having age based restricted access; for instance, voting centers, driving license centers, and traffic check posts to scan for minors, restaurants and bars to prevent under-age alcohol consumption, cinema halls to enable restricted access, or around tobacco selling vending machines. Apart from the above mentioned applications, such systems can also be deployed virtually, where access is granted based on the age; for example, in online poker rooms. 

\subsection{Literature Review: Age Estimation}
In literature, researchers have focused on estimating the age of an individual either by classification or by regression, and via hybrid methods which are a combination of both \citep{survey}. On considering it as a classification task, the problem is formulated as a $n$-class problem, where each age label is an independent class, or the age range is divided into groups, such as child, young adult, adult, and old. On the other hand, in case of regression, the age value to be estimated is considered as a series of sequential numbers. \cite{2008} have proposed a robust method to extract facial features using manifold learning and a novel method, Locally Adjusted Robust Regressor (LARR) is presented  to predict the age, where support vector regressors are explored on the learned manifold. \cite{aam} have proposed an age estimation hybrid technique based on Active Appearance Models (AAMs) and Support Vector Regression (SVR), wherein age-group specific models are utilized to perform more accurate prediction. AAMs are used to extract discriminative features from the face images based on shape as well as textural changes, while SVR is used to perform age estimation on the extracted features. \cite{pr17AgeHandcrafted} utilized several hand-crafted features such as AAM, Local Binary Patterns, Gabor Wavelets, and Local Phase Quantization, along with Support Vector Machines (SVMs) and SVRs for performing age estimation. In order to avoid overfitting while performing the task of age classification, \cite{adience} have proposed a novel approach for training SVMs using dropout. The authors show that the modified model improves the classification accuracy for the defined problem (8 age groups) as compared to the existing benchmark results. Recently, \cite{pams} have proposed to formulate the problem of age estimation by using a label distribution for a given sample rather than a single label. The label distribution represents the proportion of each age label, which is motivated by the fact that neighboring ages are highly correlated in nature, and hence cannot be described by a single age value. \cite{hooman} studied the performance of humans versus machines for age estimation and established machines' performance to supersede that of humans on the FG-Net Aging database \citep{fgnet}. Inspired by the promising performance of deep learning architectures, \cite{cvprw15} propose a lean shallow network using Convolutional Neural Networks (CNNs) for attribute prediction of unconstrained images. Recently, \cite{d2c} and \cite{pr17Age} have also demonstrated superior performance of deep learning models for performing age estimation on face images. 

In 2015, ChaLearn Looking At People (LAP) also organized a challenge as part of ICCV'15 \citep{chaLearn} to perform \textit{apparent} age estimation. Apparent age is defined as the age perceived by human beings based on the visual appearance of an individual. The dataset was created using a Facebook application to vote for the perceived age from a given image. \cite{DEX} obtained the best performance in the given challenge, wherein the authors presented a CNN based model. The VGG-16 models \citep{vgg-16} pre-trained on Image-Net dataset \citep{imagenet} are fine-tuned using the proposed IMDB-Wiki datasets for the task of apparent age estimation, and a single neuron gives the output, a whole number from [0, 100]. 
However, it is important to note that the competition is aimed at predicting the apparent or perceived age of face images, which is less relevant for monitoring cases of restricted access, as compared to the real age.

\begin{figure}
\centering
\includegraphics[width= 3.5in]{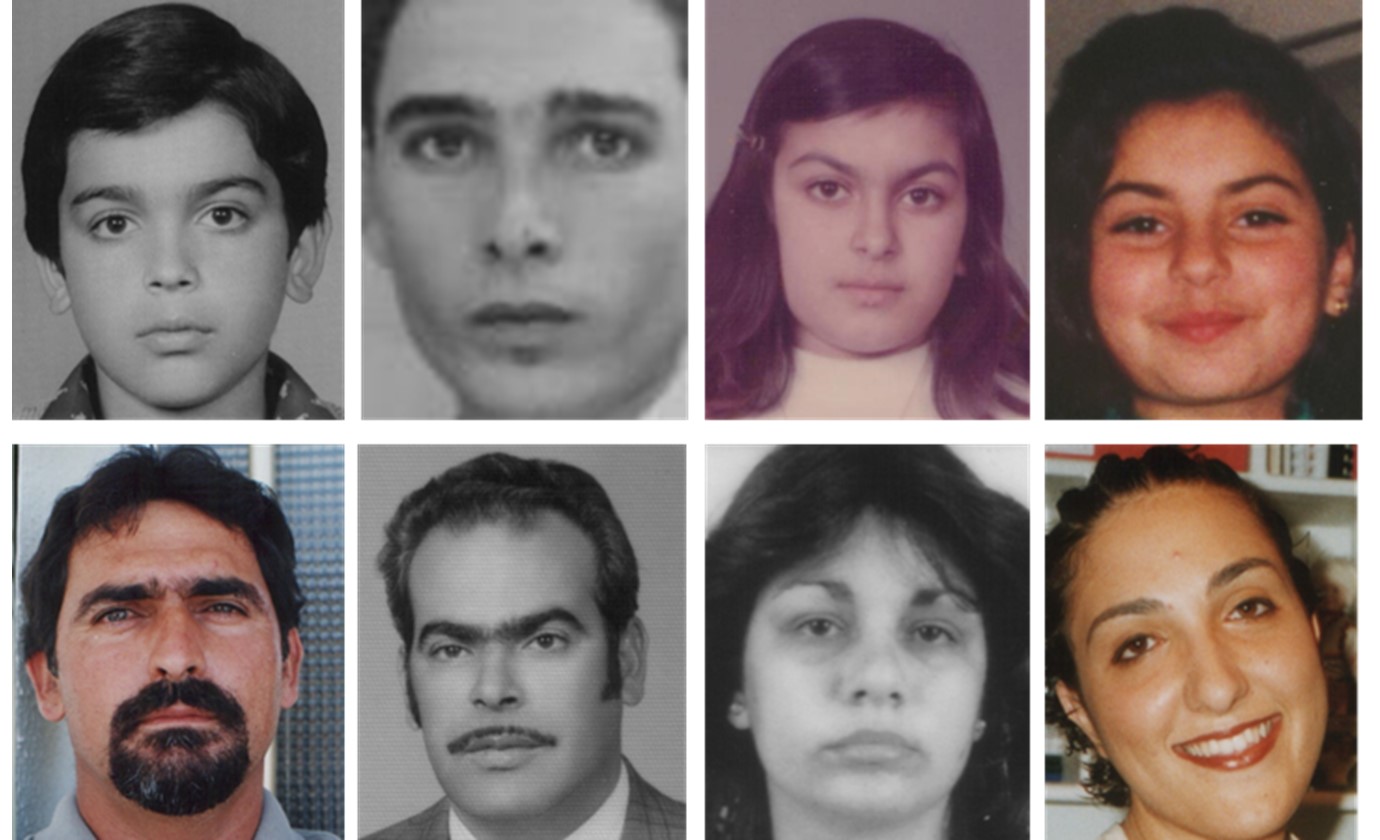}  
\caption{Demonstrating the intra-class variations in the two classes: first row represents images belonging to \textit{minor} class and second row corresponds to \textit{adult} class.}
\vspace{-10pt}
\label{fig:classImages}
\end{figure}

\subsection{Research Contributions}
In this research, we propose a deep learning based novel representation learning algorithm to determine whether the given input face image is above the age of majority or not, i.e., the image corresponds to an adult or a minor. A supervised deep learning algorithm is presented which reduces the intra-class variations at the time of feature learning. Thus, the three-fold contributions of this research are: 

\begin{itemize}
\item propose the formulation of a deep learning architecture termed as the \textit{Class Specific Mean Autoencoder}, which uses the class information of a given sample at the time of training to learn the intra-class similarity and extract similar features for samples belonging to the same class,
\item present the Multi-Resolution Face Dataset (MRFD) which contains images pertaining to 317 subjects (each having at least 12 images), out of which 307 subjects are below the threshold age of 18 years. MRFD is created due to the lack of an existing database containing images of subjects below the age of 18 years and will be made publicly available to the research community,
\item demonstrate results on the Multi-Resolution Face Dataset as part of a large combined face database of more than 13,000 images from multiple ethnicities. Results are also demonstrated on MORPH Album II database \citep{morph} containing more than 55,000 face images.
\end{itemize}

The remainder of this paper is organized as follows: Section \ref{sec:algo} provides the detailed description of the proposed algorithm. Section \ref{db} presents the datasets used, along with the experimental protocols. The results and observations are presented in Section \ref{res}, followed by the conclusions of this work.

\section{Proposed Algorithm}
\label{sec:algo}
\begin{figure}
\centering
\includegraphics[height= 1.2in]{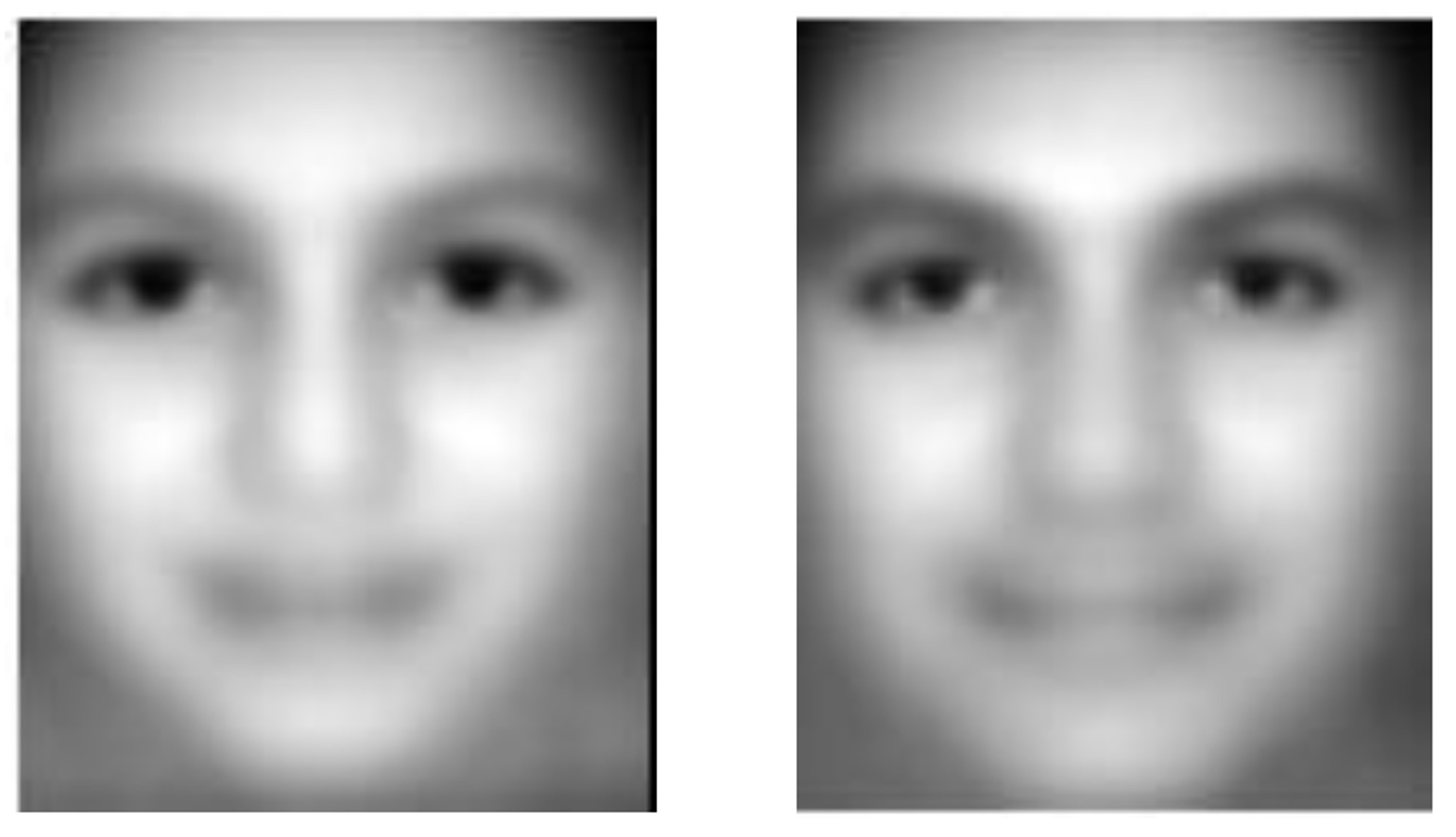}  
\vspace{-5pt}
\caption{Mean face images obtained from the images corresponding to the two age groups. The left image corresponds to the mean image of individuals below the age of majority, while the right image corresponds to the mean image of individuals above the age of majority.}
\vspace{-5pt}
\label{fig:meanImages}
\end{figure}

Deep learning architectures have been used in literature to address a large variety of tasks \citep{dlRev}. Specifically, recent models such as the FaceNet \citep{facenet}, VGG-Face \citep{vgg}, and DeepFace \citep{deepFace} have shown high performance for the task of face recognition. Models have been developed to perform automated face detection and alignment as well \citep{deepDet1, deepDet2, deepDet3}. In this work, the task of adulthood classification is addressed using a deep learning framework. As shown in Fig. \ref{fig:classImages}, the intra-class variability in both classes, minor and adult, is high. Analyzing the mean image of individual classes (as shown in Fig. \ref{fig:meanImages}) shows that the mean images of the two classes are significantly different. Based on this observation, it is our hypothesis that projecting the image features closer to the class mean can assist in learning class specific discriminative features. Therefore, in this work, we propose Class Specific Mean Autoencoder, which learns features such that the representations of the samples corresponding to a class are similar to the mean representation of the same class. Before elaborating upon the proposed model, the following subsection presents some preliminaries. 

\begin{table*}
\begin{center}
\caption{Brief literature review of autoencoder based formulations.}
\label{table:autoencoder}
\small
\noindent\makebox[\linewidth]{
\begin{tabular}{|p{2.5cm}|p{13cm}|p{1.3cm}|}
\hline
Authors & Approach & Supervised \\
\hline\hline
\cite{sdae} & Stacked Denoising Autoencoder (SDAE): Noise is added to the input data such that the learned representation is robust. & No\\
\hline
\cite{sparse} & Incorporated $\ell_1$ norm in the loss function of the autoencoder to introduce sparsity in the learned features. & No \\
\hline
\cite{Contractive} &
Contractive Autoencoder (CAE): Input space is localised by adding a penalty term which is the Jacobian of the input with respect to the hidden layer. & No \\
\hline
\cite{highContract} & Higher order Contractive autoencoder: CAE + Hessian of the output wrt the input. & No \\
\hline
\cite{contrast} & Contrastive autoencoder: A term to reduce the intra-class variations between the learned representation of samples belonging to the same class is added at the final layer. & Yes\\
\hline
\cite{generalizedAE} & Generalised Autoencoder: SDAE is modified such that the representation incorporates the structure of the dataspace as well. & No \\
\hline
\cite{smcae} & Stacked Multichannel Autoencoder: The gap between synthetic data and real data is reduced by learning a mapping between the two. & No \\
\hline
\cite{gao2015} & Inspired from SDAE, an identification specific model is proposed, where the probe image is treated as the noisy input while the gallery images are treated as the clean input. & Yes \\
\hline
\cite{transferAE} & A two layer model is proposed wherein, a representation is learned in the first layer, and the class label is encoded in the second layer. & Yes \\
\hline
\cite{majumdar2016} & A joint sparsity (using $\ell_{2,1}$) promoting supervision penalty term is added to the loss function of SDAE. & Yes \\
\hline
\cite{relational} & A relational term, which aims to model the relationship between the input data is added to the loss function. & No \\
\hline
Proposed (2018) & Class Specific Mean Autoencoder: utilizes class mean to learn discriminative features. & Yes \\
\hline
\end{tabular}
}
\label{tab:lit}
\end{center}
\vspace{-15pt}
\end{table*}

\subsection{Preliminaries: Supervised Autoencoders}

Several researchers have proposed modifications to the traditional autoencoder architecture. Table \ref{tab:lit} provides a summary of these architectures. Most of these are unsupervised in nature, however, researchers have proposed supervised architectures that leverage the availability of labeled data as well. In this section, we briefly present the original formulation of autoencoder followed by discussing the existing supervised architectures.

For a given input $x$, the loss function of a single layer traditional autoencoder \citep{ae} is given as follows: 
\begin{equation} \label{eqLossAe}
\argmin_{\textit{$\mathbf{W}_e,\mathbf{W}_d$}}  \left \|x -  \textbf{W}_d\phi(\textbf{W}_ex) \right \|_{2}^{2} 
\end{equation}
\noindent where, $\mathbf{W}_e$ and $\mathbf{W}_d$ are the respective encoding and decoding weights of the autoencoder, and $\phi$ corresponds to an activation function, generally incorporated for introducing non-linearity in the model. Common examples of activation functions are $sigmoid$ and $tanh$.  An autoencoder learns features $\left( f_x = \phi(\mathbf{W}_ex)\right)$ of the given input ${x}$, such that the error between the original sample and it's reconstruction ($\mathbf{W}_df_x)$ is minimized. For a $k$ layered autoencoder, having encoding weights as $\mathbf{W}_e^1, \mathbf{W}_e^2, ..., \mathbf{W}_e^k$, and decoding weights as $\mathbf{W}_d^1, \mathbf{W}_d^2, ..., \mathbf{W}_d^k$, the loss function of Eq. \ref{eqLossAe} is modified as follows:
\begin{equation} \label{eqLossMulti}
\argmin_{\textit{$\substack{\mathbf{W}_e^1, ..., \mathbf{W}_e^k,  \mathbf{W}_d^1, ..., \mathbf{W}_d^k}$}}  
\left \|x -  b\circ a\ (x)\right \|_{2}^{2}  
\end{equation}
\noindent where, $a({x}) =\phi(\mathbf{W}_e^k (\phi(\mathbf{W}_e^{k-1}\dots(\phi(\mathbf{W}_e^1x)))))$ refers to the encoding function, and $b\ (x) = \mathbf{W}_d^1(\mathbf{W}_d^{2} \dots (\mathbf{W}_d^kx))$ corresponds to the decoding function. The first and the last layers correspond to the input and output layers respectively, while the remaining layers are often termed as the hidden layers. 

In literature, researchers have incorporated class information in the traditional formulation of an autoencoder in order to facilitate supervision. \cite{gao2015} modify the denoising autoencoder \citep{sdae} to learn supervised image representations in order to optimize the identification performance. At the time of training, for a given subject, the probe image is the input to the autoencoder (analogous to the noisy input), and the gallery image of the subject (analogous to the clean image) is the target image used for computing the reconstruction error, as in the case of a denoising autoencoder. A similarity preservation term is added to the loss function such that the samples belonging to the same class have a similar representation. Given probe and gallery images of class $i$, each probe image is represented using {${{x_n}_i}$} and its corresponding gallery images are represented using ${x_i}$. The loss function for the supervised autoencoder is as follows:
\begin{equation} \label{Supervised}
\begin{gathered}
\argmin_{\textit{$\substack{\mathbf{W}_e^1, ..., \mathbf{W}_e^k, \\ \mathbf{W}_d^1, ..., \mathbf{W}_d^k}$}} \frac{1}{N} \sum_\textit{i} \bigg( \left \|{x_i} - b \circ a({{{x_n}_i}}) \right \|_{2}^{2} + \lambda \left \|a({x_i}) - a({{{x_n}_i}}) \right \|_{2}^{2}\bigg) \\
+ \alpha \bigg( KL ( \rho_ x \| \rho_o ) + KL ( \rho_{x_n} \| \rho_o) \bigg) \\
where, \rho_x = \frac{1}{N} \sum_i \frac{1}{2} \Big( a({x_i}) + 1 \Big) \;\; \; \;  and \\
\rho_{{{x_n}_i}} = \frac{1}{N} \sum_i \frac{1}{2} \Big( a({{x_n}_i}) + 1\Big) 
\end{gathered}
\end{equation}
\noindent here, the first term corresponds to the reconstruction error, second is the similarity preservation term, and the remaining two terms correspond to the Kullback Leibler (KL) divergence \citep{kullback} to introduce sparsity in the hidden layers.

Contrastive Autoencoder (CsAE) proposed by \cite{contrast}, is another variant of supervised autoencoder which uses the class label information during training. The loss function of the model is the difference between the output of two sub-autoencoders trained simultaneously on samples belonging to the same class, along with the loss function of each sub-autoencoder. The equation for the same is given as:
\begin{equation}\label{eqLossCsAE}
\begin{gathered}
\argmin_{\textit{$\substack{\mathbf{W}_e^1, ..., \mathbf{W}_e^k, \\ \mathbf{W}_d^1, ..., \mathbf{W}_d^k}$}} \lambda(\left \|{x_1} - b \circ a({x_1}) \right \|_{2}^{2} + \left \|{x_2} - b \circ a({x_2}) \right \|_{2}^{2}) \\
+ (1-\lambda) \left \|O_k({x_1}) - O_k({x_2}) \right \|_{2}^{2}. \\
\end{gathered}
\end{equation}
where, ${x_1}$ and ${x_2}$ represent two different input samples belonging to the same class. For each sub-autoencoder, $a(x) = \phi(\mathbf{W}^{k}_e \phi(\mathbf{W}^{k-1}_e ... \phi(\mathbf{W}^1_e(x)))$ and $b(x) = \mathbf{W}^1_d (\mathbf{W}^2_d... \mathbf{W}^k_d (x))$, where $\mathbf{W}^i_e$ and $\mathbf{W}^i_d$ refer to the encoding and decoding weights of the $i^{th}$ layer, and \textit{$O_k(x)$} is the output of the \textit{$k^{th}$} layer.

Recently, \cite{majumdar2016} present a class sparsity based supervised encoding algorithm wherein a joint-sparsity promoting $l_{2,1}-$norm supervision penalty is added to the loss function. For samples $\mathbf{X}$, belonging to total $C$ classes, the modified algorithm is presented as:
\begin{equation} \label{pami}
\argmin_{\substack{\mathbf{W}_e^1, ..., \mathbf{W}_e^k, \mathbf{W}_d^1, ..., \mathbf{W}_d^k}} \left \|\mathbf{X} - b\circ a(\mathbf{X}) \right \|_{F}^{2} + \lambda \sum_{c=1}^{C}\left \| \mathbf{W}_e\mathbf{X}_c\right \|_{2,1}
\end{equation}
where, $\mathbf{X}_c$ refers to the samples belonging to class $c$. The regularization term enforces same sparsity signature across each class, which leads to similar representations of samples from a given class.

\begin{figure}
\centering
\includegraphics[width= 3.5in]{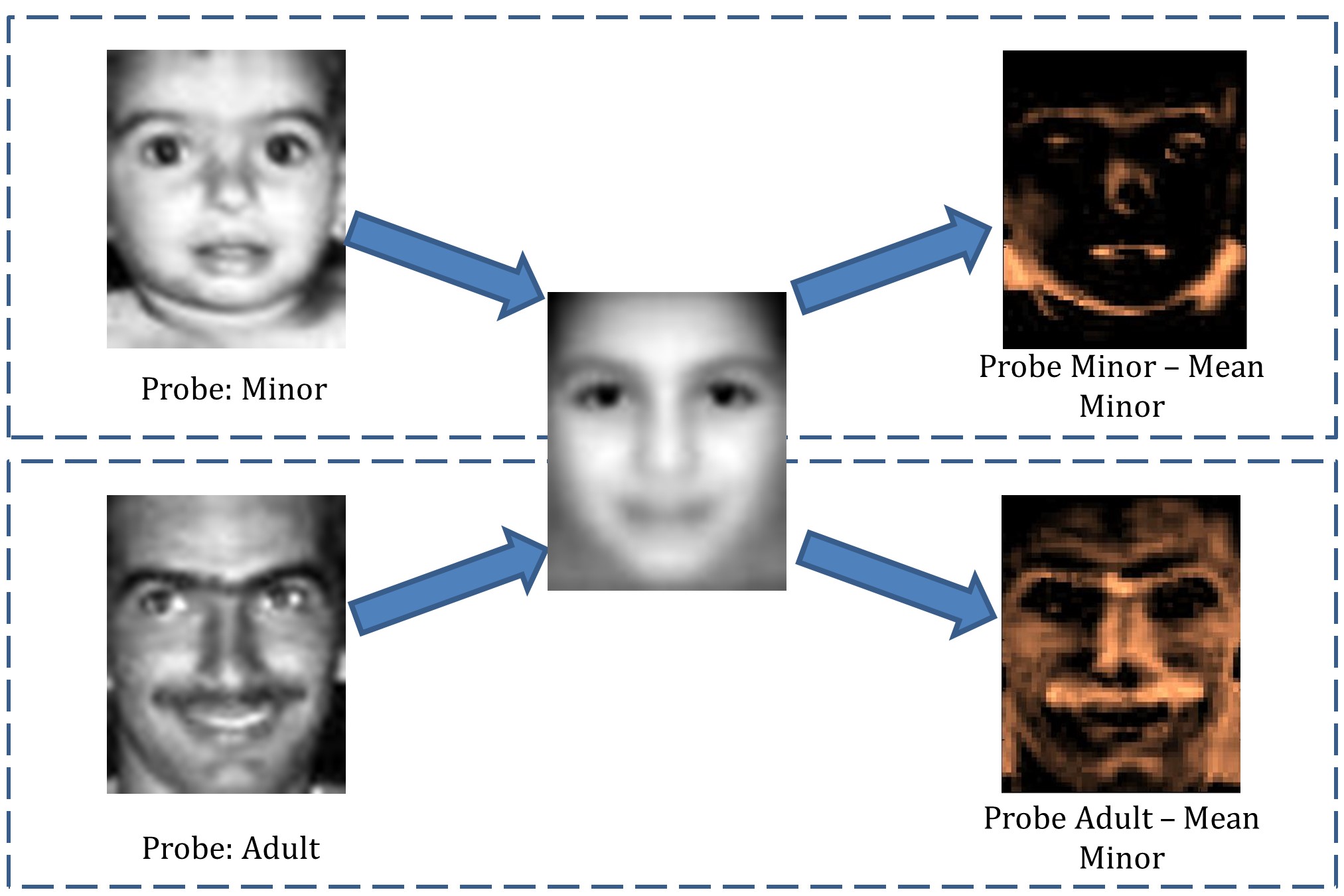} 
\vspace{-15pt}
\caption{For a two class problem consisting of \textit{adults} and \textit{minors}, and two probe images, the figure depicts the difference of each probe with respect to the mean minor image. It can be observed that the difference of the minor probe image from the mean minor image is significantly less than the difference of the adult probe image with the mean minor image. This motivates the use of class specific mean feature vectors for incorporating supervision in the feature learning process.}
\label{fig:mean}
\vspace{-5pt}
\end{figure}

\begin{figure*}[t]
\centering
\includegraphics[width= 7.2in]{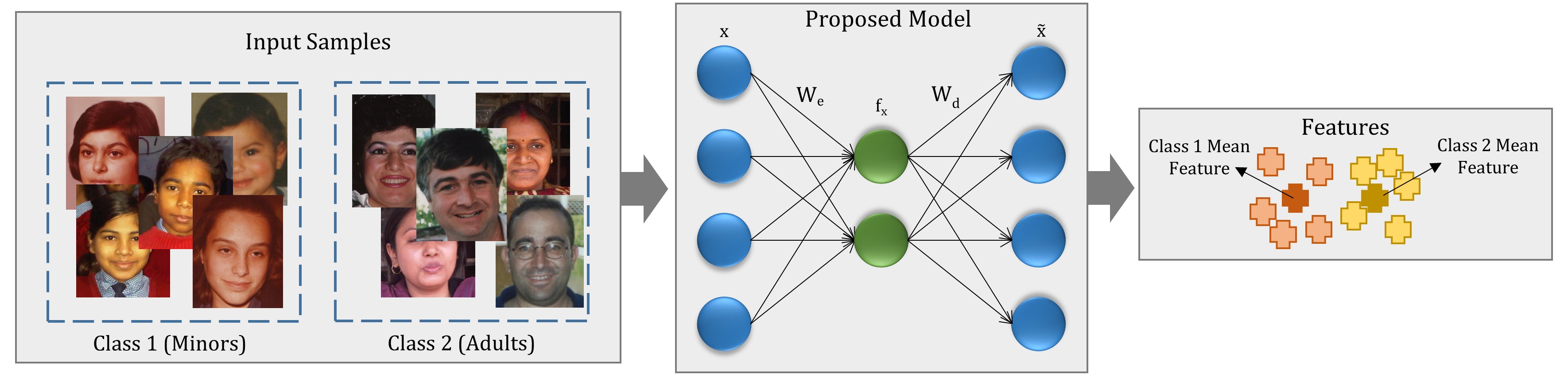} 
\vspace{-15pt}
\caption{Proposed Class Specific Mean Autoencoder. ${x}$ and ${\tilde{x}}$ represent the input and the reconstructed sample respectively, $\mathbf{W}_e$ and $\mathbf{W}_d$ denote the encoding and decoding weights, and ${f_x}$ corresponds to the learned feature vector. }
\label{fig:proposed}
\vspace{-10pt}
\end{figure*}

\subsection{Proposed Class Specific Mean Autoencoder}
While all the above techniques incorporate supervision into an otherwise unsupervised model, the proposed architecture incorporates the mean feature of each class into the feature learning process as well. The key motivation behind the proposed algorithm is illustrated in Fig. \ref{fig:mean}. In this example, with faces as input, \textit{adult} and \textit{minor} as the two classes, the mean adult image and mean minor image are computed using the training samples. For a given probe face image, computing the $l_2$ distance with respect to the mean minor image provides the similarity of the sample with the minor class. It can be observed that the difference between a \textit{minor} probe image and the mean \textit{minor} image is lower as compared to the difference between an \textit{adult} probe image and the mean \textit{minor} image. This example shows that if the intra-class variations are encoded, it may help in learning class-specific features. Inspired from this observation, in this research, we present a novel formulation of Class Specific Mean Autoencoder.

In the proposed formulation, the loss function of an autoencoder \citep{ae} is updated by introducing class specific information. For simplicity and clarity, Eq. \ref{eqLossAe} is repeated as follows:
\begin{equation} \label{eqLossAeRep}
{
\argmin_{\textit{$\mathbf{W}_e,\mathbf{W}_d$}}  \left \|x -  \mathbf{W}_d\phi(\mathbf{W}_ex) \right \|_{2}^{2} 
}
\end{equation}

For an input sample $x_c$, belonging to class $c$, the feature vector $f_{x_c}$ is defined as follows:
\begin{equation} \label{eqFeat}
f_{x_c} = \phi(\mathbf{W}_ex_c)
\end{equation}
\noindent The mean feature vector pertaining to the $c^{th}$ class is defined as:
\begin{equation} \label{eqMean}
m_c = \mu(\phi(\mathbf{W}_e\mathbf{X}_c))
\end{equation}
\noindent where, $\mu$ represents the mean operator, and $\mathbf{X}_c$ represents all the training samples belonging to class $c$.

As discussed earlier in this section, we postulate that encoding the difference between the feature of a sample and the mean sample of the same class can help in encoding class-specific features. In other words, the feature of a particular class is brought similar/closer to the \textit{mean} feature of that class. To encode this information, Eqs. \ref{eqFeat} and \ref{eqMean} are utilized to form the following optimization constraint:
\begin{equation} \label{eqReg}
\left \|f_{x_c} - m_c  \right \|_{2}^{2} 
\end{equation}

The above equation is incorporated into an autoencoder to create Class Specific Mean Autoencoder as follows:
\begin{equation} \label{eqLossProp}
\argmin_{\textit{$\mathbf{W}_e, \mathbf{W}_d$}}  \left \|x_c -  \mathbf{W}_d\phi(\mathbf{W}_ex_c) \right \|_{2}^{2}  + \lambda \left \|f_{x_c} - m_c \right \|_{2}^{2} 
\end{equation}
where, $\lambda$ is the regularization constant. The proposed Class Specific Mean Autoencoder learns the weight parameters such that the features of a particular class are \textit{grouped} together. Expanding Eq. \ref{eqLossProp}, we obtain:
\begin{equation} \label{eqLossPropExp}
\begin{gathered}
\argmin_{\textit{$\mathbf{W}_e, \mathbf{W}_d$}}  \left \|x_c -  \mathbf{W}_d\phi(\mathbf{W}_ex_c) \right \|_{2}^{2}  + \\ \hspace{2cm} \lambda \left \|\phi(\mathbf{W}_ex_c) - \mu(\phi(\mathbf{W}_e\mathbf{X}_c)) \right \|_{2}^{2} 
\end{gathered}
\end{equation}
The updated loss function of Eq. \ref{eqLossPropExp} ensures that the learned feature for a sample is close to the mean representation of its class, while being representative of the input sample as well. The second term is added for supervised regularization and can be viewed as: 
\begin{equation} \label{eqReg}
E =  \left \|f_{x_c} - t \right \|_{2}^{2} 
\end{equation}
for a given expected target $t$ and obtained output $f_{x_c}$. The above equation draws a direct parallel with Eq. \ref{eqLossAe}, where the expected target is $x$, and the obtained output is ($\mathbf{W}_d\phi(\mathbf{W}_ex)$). Similar to the update rule for Eq. \ref{eqLossAe}, the update rule for the above regularization term for $j^{th}$ expected ($t_j$) and obtained ($o_j$) output, with respect to weight $w_{e_{i,j}}$, can be written as:
\begin{equation} \label{eqRegDiff}
\frac{\partial \mathbf{E}}{\partial \mathbf{W}_{e_{i,j}}} = \frac{1}{2}*(o_j - t_j)* \frac{\partial o_j}{\partial \mathbf{W}_{e_{i,j}}} 
\end{equation}
Similar to the gradient descent backpropogation applied to Eq. \ref{eqLossAe}, the Class Specific Mean Autoencoder is solved iteratively via the above update rule till convergence.  

For a $k$ layered Class Specific Mean Autoencoder, having encoding weights as $\mathbf{W}_e^1, \mathbf{W}_e^2, ..., \mathbf{W}_e^k$, and decoding weights as $\mathbf{W}_d^1, \mathbf{W}_d^2, ..., \mathbf{W}_d^k$, the loss function of Eq. \ref{eqLossProp} can be modified as:
\begin{equation} \label{eqLossPropMulti}
\argmin_{\textit{$\substack{\mathbf{W}_e^1, ..., \mathbf{W}_e^k, \\ \mathbf{W}_d^1, ..., \mathbf{W}_d^k}$}}  
\left \|x_c -  b\circ a(x_c) \right \|_{2}^{2}  + \sum_{i = 1}^{i=k} \lambda_i \left \|f_{x_c}^i - m_c^i \right \|_{2}^{2} 
\end{equation}
where, $a(x) =\phi(\mathbf{W}_e^k (\phi(\mathbf{W}_e^{k-1}\dots(\phi(\mathbf{W}_e^1x)))))$ is the encoding function, and $b(x) = \mathbf{W}_d^1(\mathbf{W}_d^{2} \dots (\mathbf{W}_d^kx))$ corresponds to the decoding function, and $f_{x_c}^k$ and $m_c^k$ are defined as:
\begin{equation} \label{eqFeatLayer}
f_{x_c}^k = \phi(\mathbf{W}_e^k (\phi(\mathbf{W}_e^{k-1}\dots(\phi(\mathbf{W}_e^1x_c)))))
\end{equation}
\begin{equation} \label{eqMeanLayer}
m_c^k = \mu(\phi(\mathbf{W}_e^k (\phi(\mathbf{W}_e^{k-1}\dots(\phi(\mathbf{W}_e^1\mathbf{X}_c))))))
\end{equation}
 
Owing to the large number of parameters involved, the optimization of the above model is performed via the greedy layer by layer approach \citep{greedy}. At the time of testing, the learned encoding weights ($\mathbf{W}_e^1, \mathbf{W}_e^2, \dots \mathbf{W}_e^k$) are used to calculate the feature vector for a given sample, which is then provided as input to a classifier. Fig. \ref{fig:proposed} presents a pictorial representation of the proposed algorithm, for a two class problem. 

\begin{algorithm} 
    \SetKwInOut{Input}{Input}
    \SetKwInOut{Output}{Output}
    \Input{Training images of minor ($\mathbf{X}_{minor}$) and adult ($\mathbf{X}_{adult}$) classes, iter = 0, maxIter. }
    \Output{Encoding and decoding weights: $\mathbf{W}_e$, $\mathbf{W}_d$.}
    Initialize $\mathbf{W}_e$ and $\mathbf{W}_d$\;
    \While{iter $<$ maxIter}
    {
    Compute mean adult feature (${m}^{iter}_{adult}$) using Eq. \ref{eqMean} \;
    Compute mean minor feature (${m}^{iter}_{minor}$) using Eq. \ref{eqMean} \;
    \ForEach{$x_{minor} \in \mathbf{X}_{minor}$}{
    Minimize Eq. \ref{eqLossProp} using $x_{minor}$ and ${m}^{iter}_{minor}$\;
    }
    \ForEach{$x_{adult} \in \mathbf{X}_{adult}$}{
    Minimize Eq. \ref{eqLossProp} using $x_{adult}$ and ${m}^{iter}_{adult}$\;
    }
    iter++\;
    }

    \caption{Training Single Layer Class Specific Mean Autoencoder for Adulthood Prediction}
    \label{algo}
\end{algorithm}

\begin{table*}[h]
\begin{center}
\caption{Summarizing the dataset description and experimental protocol.}
\vspace{5pt}
\label{tab:protocol}
\begin{tabular}{|l|c|c|c|c|c|}
\hline
\multicolumn{1}{|l|}{\multirow{2}{*}{\textbf{Database}}} & \multicolumn{1}{c|}{\multirow{2}{*}{\textbf{Total Number of Images}}} & \multicolumn{2}{c|}{\textbf{Number of Images of}}                  & \multicolumn{2}{c|}{\textbf{Number of Images in}}                       \\ \cline{3-6}
&   & \multicolumn{1}{c|}{\textbf{Minors}} & \multicolumn{1}{c|}{\textbf{Adults}} & \multicolumn{1}{c|}{\textbf{Train Set}} & \multicolumn{1}{c|}{\textbf{Test Set}}  \\ \cline{1-6}\hline
\hline
MORPH Album-II Dataset & 55,132 & 3,330 & 51,802 & 4,662 & 50,470 \\
\hline
Multi-Ethnicity Dataset & 13,133 & 8,574 & 4,559 & 6,276 & 6,857 \\
\hline
\end{tabular}
\end{center}
\vspace{-15pt}
\end{table*}

\subsection{Predicting Adulthood using Class Specific Mean Autoencoder}
The proposed Class Specific Mean Autoencoder is used to address the problem of classification of face images into adults (18 years of age or more) or minors (less than $18$ years of age). The proposed model is used for feature extraction, which is then followed by a Neural Network for classification. The algorithm is summarized in Algorithm \ref{algo}.

\begin{figure}[h]
\centering
\subfloat[Multi-Ethnicity Dataset]{\includegraphics[width = 3.3in]{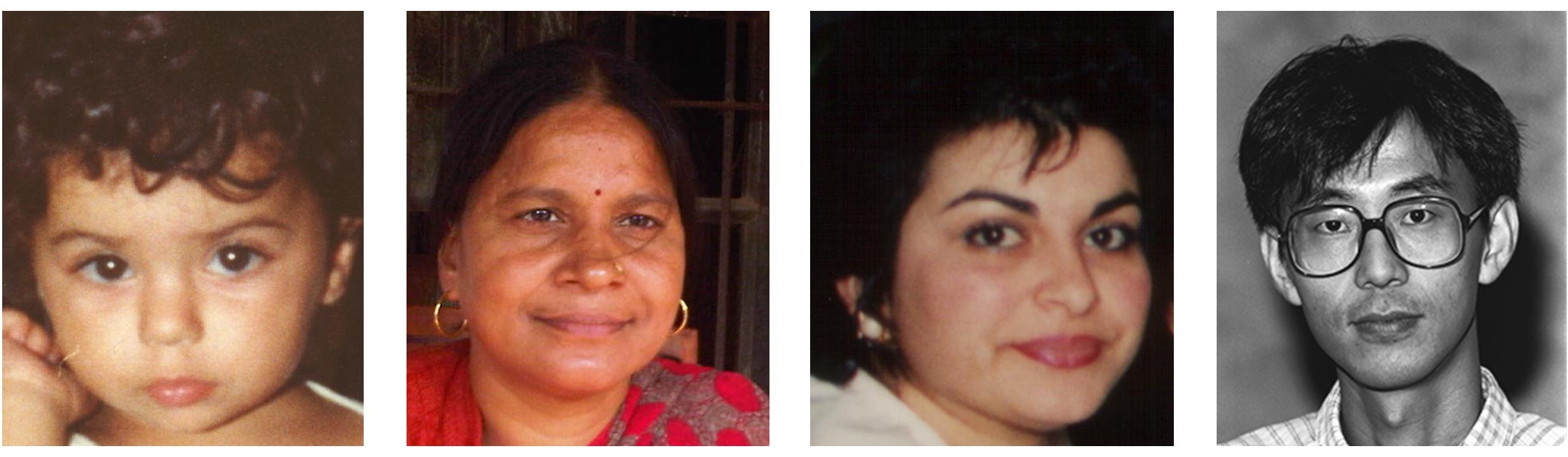}} \\ 
\vspace{-5pt}
\subfloat[MORPH Album-II Dataset]{\includegraphics[width= 3.3in]{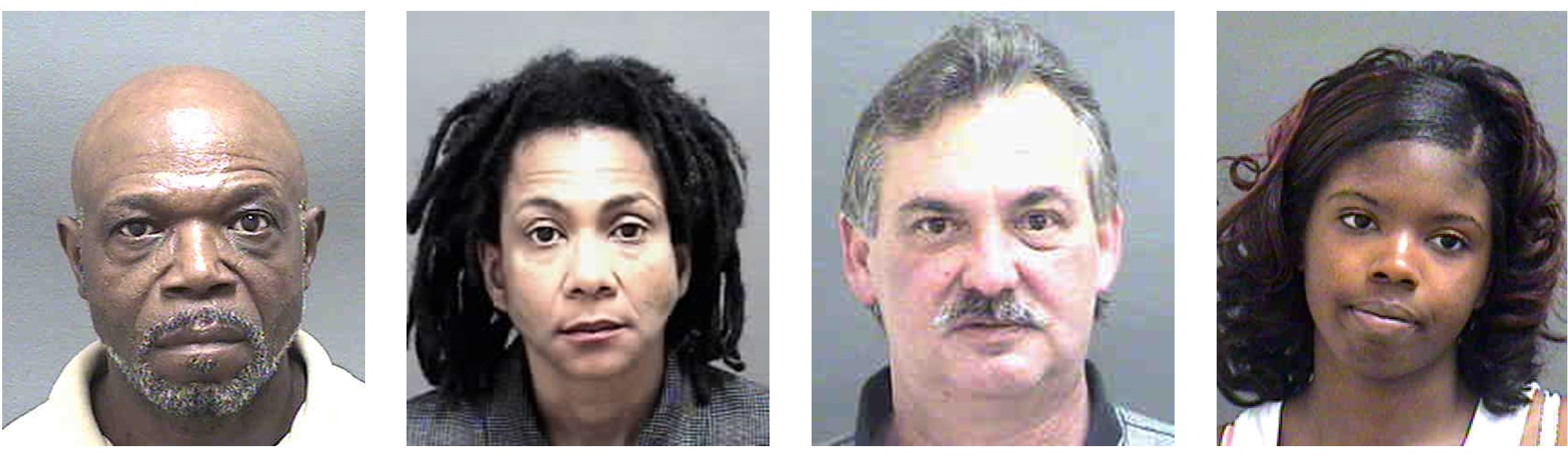}} 
\vspace{-5pt}
\caption{Sample images from the datasets used for experimental evaluation.}
\label{fig:sampleDb}
\vspace{-5pt}
\end{figure}

\begin{figure} [h]
\centering
\includegraphics[width= 2.8in]{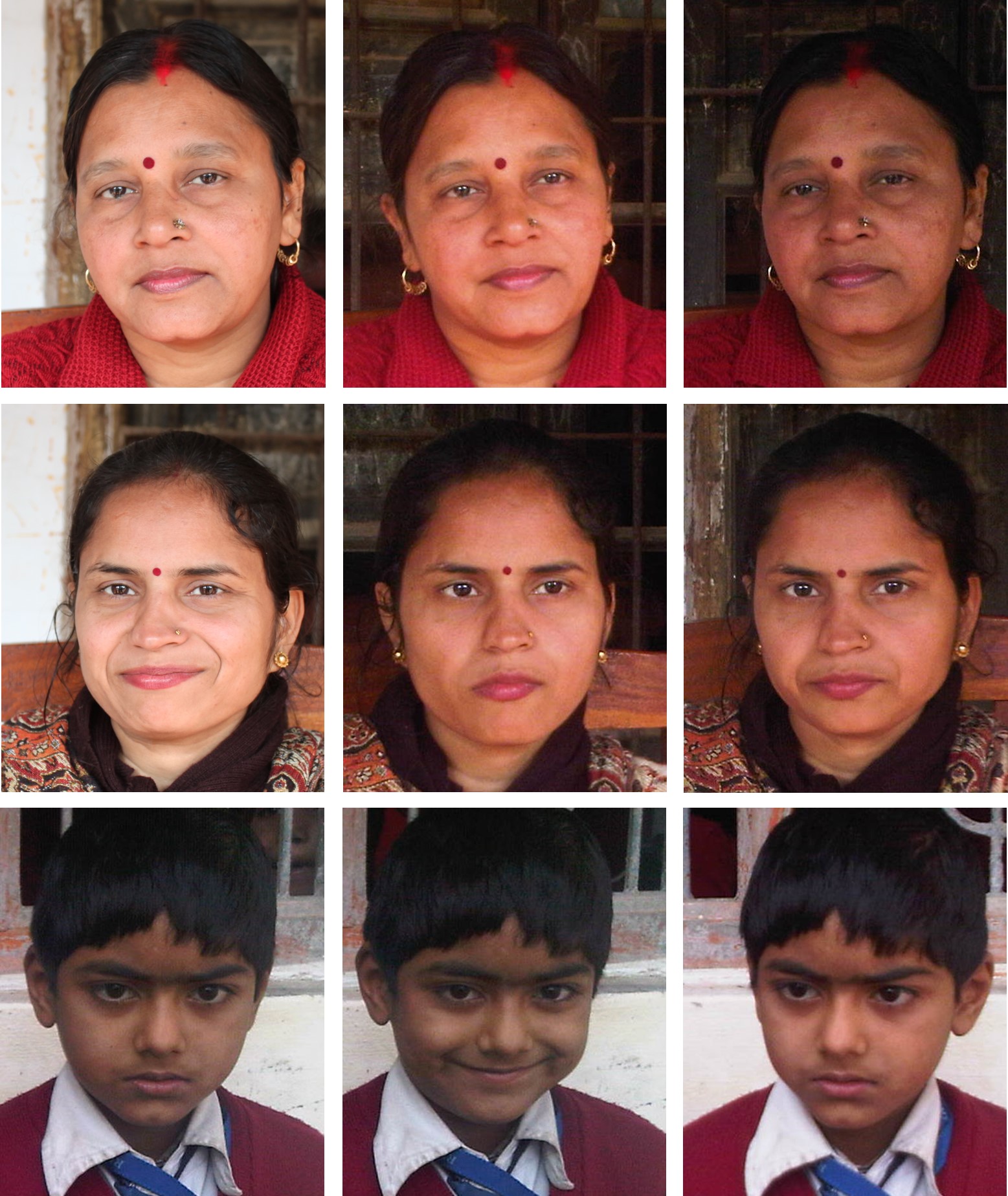} 
\caption{Sample images from the proposed Multi-Resolution Face dataset. }
\label{fig:multiResSamples}
\vspace{-10pt}
\end{figure}

\section{Datasets and Experimental Details}
\label{db}
Given a face image, predicting whether the individual is an adult or not, can be modeled as a two class classification problem: individuals below the age of 18 years are referred as \textit{minors}, while individuals of age equal to or greater than 18 years are referred as \textit{adults}. Details regarding the datasets used, and the experimental protocols are as follows:

\subsection{Datasets Used}
Experiments are performed on two datasets: (i) Multi-Ethnicity dataset and (ii) MORPH Album-II dataset. Fig. \ref{fig:sampleDb} shows some sample images from both the datasets. Details about each are given below:

\subsubsection{Multi-Ethnicity Dataset}
Since the existing datasets containing images of minors and adults contain very limited variations with respect to ethnicity, pose, and expression, along with very few samples below the age of 16, we propose the Multi-Ethnicity Dataset. Multi-Ethnicity dataset consists 13,133 face images combined from
\begin{itemize}
\item Proposed Multi-Resolution Face Dataset containing 4,019 face images, \vspace{-5pt}
\item Heterogeneous Dataset containing 8,112 face images \citep{het}, and \vspace{-5pt}
\item FG-Net Aging Dataset containing 1,002 face images \citep{fgnet}.
\end{itemize}

The dataset contains variations across ethnicity, gender, resolution, illumination, as well as minute pose and expression. Due to the lack of datasets containing face images of both adults and minors, we have also created \textit{Multi-Resolution Face Dataset (MRFD)} consisting of 4,019 face images of minors and adults of two resolutions with slight variations in pose, expression, and illumination. The Multi-Resolution Face Dataset\footnote{The dataset will be made publicly available for academic research via http://iab-rubric.org/resources.html} consists of 4,019 Indian face images of 317 subjects captured in outdoor, as well as indoor environment. The dataset consists of images of 307 minors (3,896 images) and 10 adults. Images have been captured from two smartphones (with 3.1 MP camera) with resulting face size of $360\times420$ pixels, and a high resolution hand-held Canon digital camera with resulting face images of dimension $560\times680$. Each subject has at least 12 near-frontal, well-illuminated images (at least 4 from each camera source). The dataset contains variations in age, ranging from toddlers to adults of around 50 years. The subjects were only asked to look at the camera, without any instructions for pose or expression, which resulted in images with varying head movement and expression. This is the first dataset containing such large number of minor images which would help in facilitating research on minor face images as well. Fig. \ref{fig:multiResSamples} presents some sample images from the MRF dataset illustrating variations in age, illumination, and expression. 

\subsubsection{MORPH Album-II Dataset}
Craniofacial Longitudinal Morphological Face (MORPH) dataset \citep{morph} consists of two albums: Album-I contains scanned digital face images, while Album-II contains longitudinal digital face images captured over several years. A subset of Album-II containing 55,134 images of 13,000 subjects is made available for academic researchers, which has been used for experimental analysis in this research. The dataset contains images of subjects between the age range of 16 to 77 years, and also provides metadata for race, gender, date of birth, and date of acquisition. 
 
\begin{figure*}
\centering
\subfloat[Multi-Ethnicity Dataset]{\includegraphics[height= 3in]{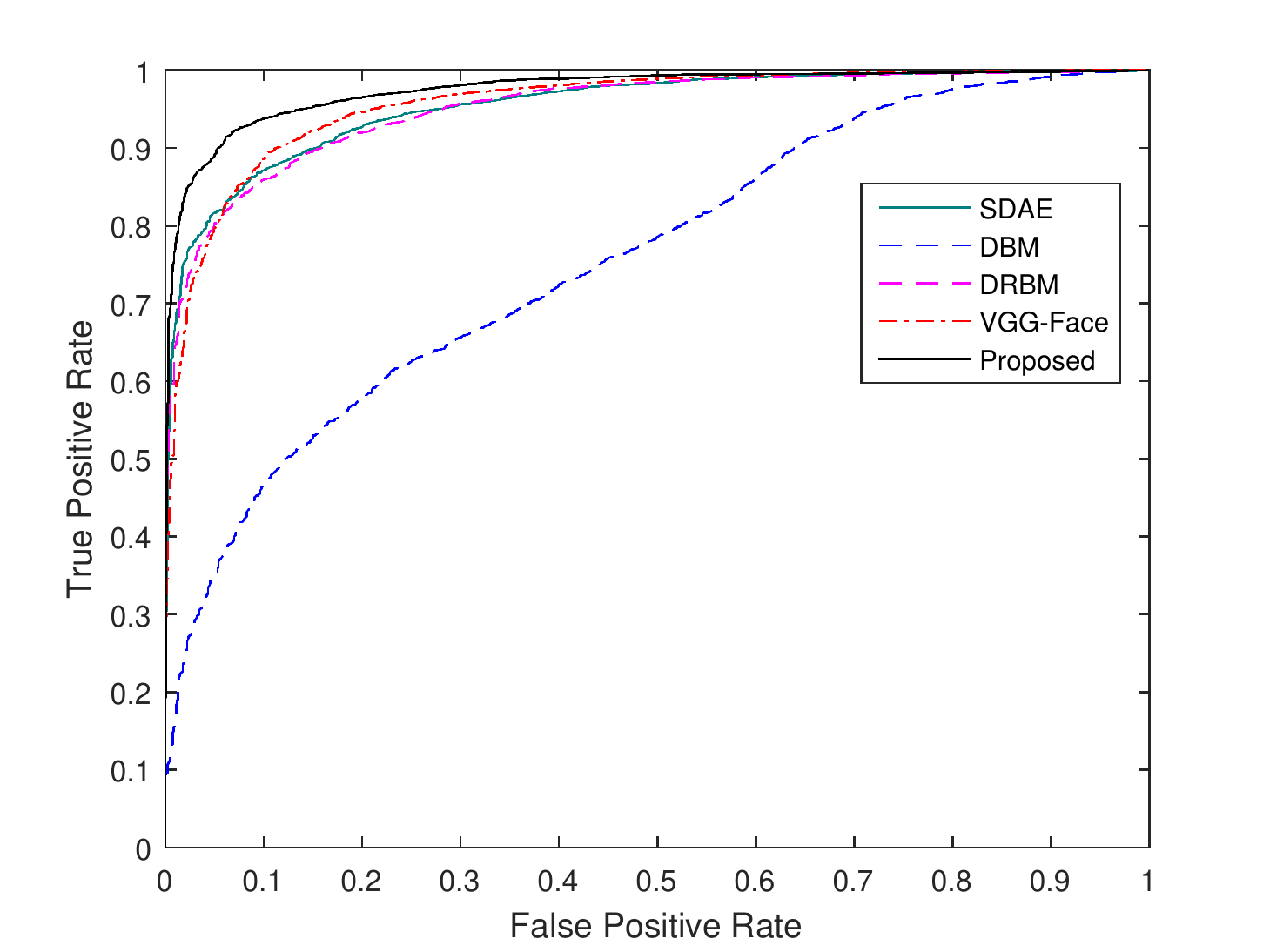}} \hspace*{-2em}
\subfloat[MORPH Album-II Dataset]{\includegraphics[height= 3in]{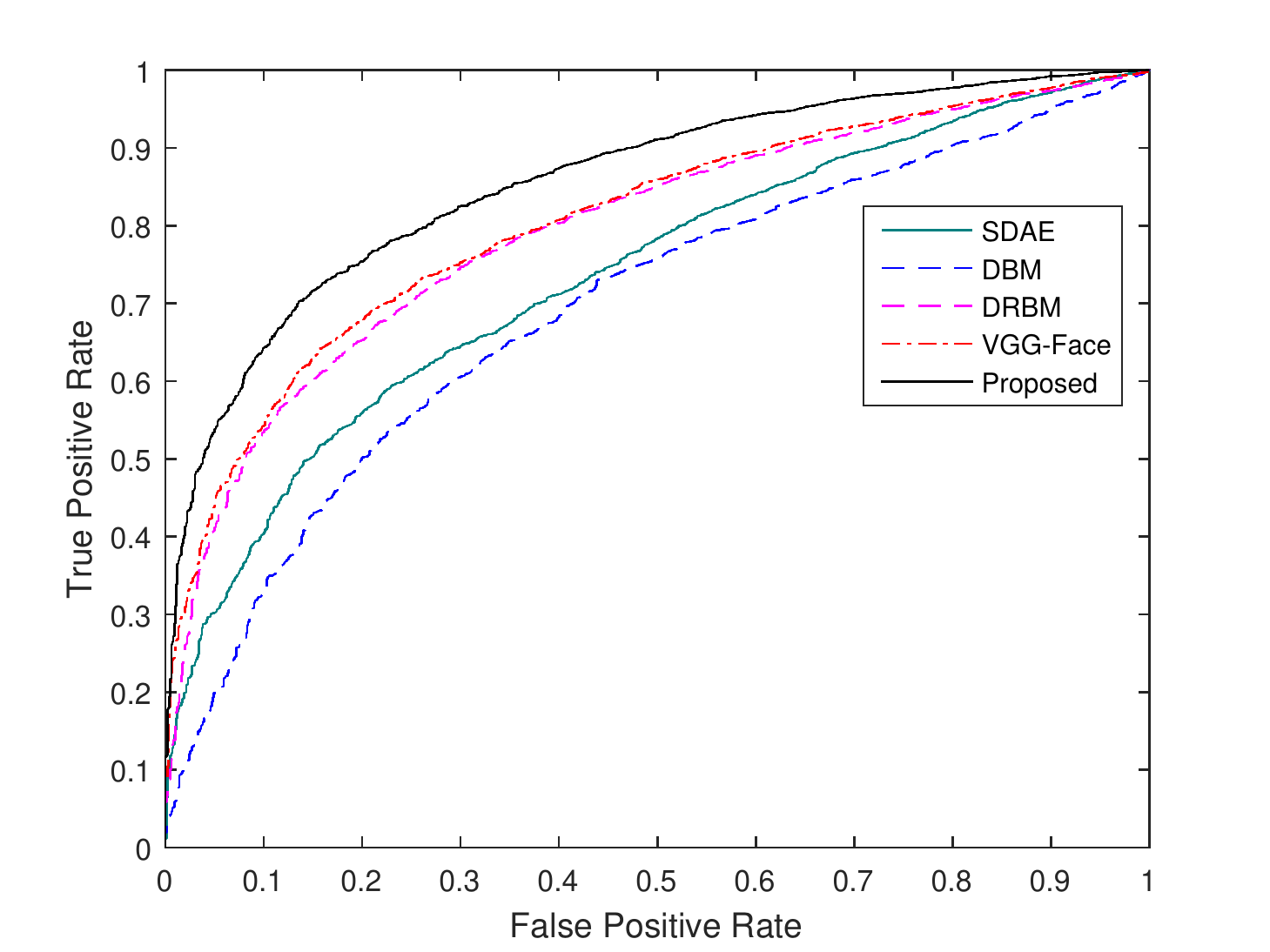}}   
\vspace{-5pt}
\caption{Receiver Operating Characteristic (ROC) curves obtained for categorizing whether a given face image is of an adult or not.}
\label{fig:roc}
\vspace{-5pt}
\end{figure*}

\subsection{Experimental Protocol}
Unseen training and testing partitions are created for both the datasets. For training, equal number of samples from both the classes are used, which is defined by the class with lesser number of samples. 70\% of the samples corresponding to the minor class and equal number of images from adult class are used for training, while the remaining data is used for testing. For the MORPH dataset, this results in the training and testing sets of size 4,662 and 50,470 images respectively. Similarly, for the Multi-Ethnicity dataset, 6,276 images are selected for training with the constraint that equal number of samples are selected from both the classes. The remaining face images constitute the test set. Details of data partitioning are documented in Table \ref{tab:protocol}.

To showcase the efficacy of the proposed algorithm, comparison has been drawn with other deep learning based feature extractors; namely, Stacked Denoising Autoencoder (SDAE) \citep{sdae}, Deep Boltzmann Machine (DBM) \citep{rbm}, and Discriminative Restricted Boltzmann Machine (DRBM) \citep{drbm}. Comparison has also been drawn with VGG-Face descriptor \citep{vgg}, which is one of the state-of-the-art deep learning based feature extractor. Features extracted from these models are provided as input to a Neural Network for classification. A CNN based Commercial-Off-The-Shelf (COTS) system, Face++ \citep{faceplusplus}, has also been used to compare the performance of the proposed model. Since there does not exist any COTS for the task of adulthood prediction, Face++ is used to estimate the age of the given face image, which is then utilized to classify the input as an adult or a minor. In order to analyze the statistical significance of the results obtained by the proposed model, McNemar test \citep{mcnemar} has been performed. Given the classification results obtained from two models, McNemar test predicts whether the performance of both the models is statistically different or not. For every comparison of the proposed Class Specific Mean Autoencoder with an existing architecture, a $p$-value is reported. A smaller $p$-value corresponds to a higher confidence level of statistical difference. In this research, all claims of statistical significance have been made at a confidence level of 95\%.

\subsection{Implementation Details}
For all the experiments, face detection is performed on all images using Viola Jones Face Detector \citep{viola}, following which the images are geometrically normalized and resized to a fixed size. A Class Specific Mean Autoencoder of dimensions [$m$, $m$] is learned, where $m$ is the size of the image. Following this, a neural network of dimension $[\frac{m}{4}, \frac{m}{8}]$ is trained for classification. $sigmoid$ activation function is used at the hidden layers. Models are trained for 100 epochs with a learning rate of 0.01. We have followed the best practices used for setting the parameters and architecture for deep learning \citep{jmlr10}. For existing algorithms, in order to maintain consistency, a two layer architecture is utilized for the feature extractor and neural network. 

\section{Experimental Results and Observations}
\label{res}
Owing to the large class imbalance in the test samples, mean class-wise accuracy has been reported for all the experiments. The formula used for calculating the accuracy is as follows: 
\begin{equation}
Accuracy = \frac{Accuracy_{Minor} + Accuracy_{Adult}}{2}
\end{equation}
where, $Accuracy_{Minor}$ and $Accuracy_{Adult}$ correspond to the accuracies obtained for minor and adult classification, respectively by a particular model. 

\subsection{Results on Multi-Ethnicity Dataset}
Table \ref{tab:mixed} presents the classification accuracies of the proposed model along with other existing architectures on the Multi-Ethnicity Dataset. Fig. \ref{fig:roc} also presents the Receiver Operative Characteristic (ROC) curves obtained for the experiments. It is observed that the proposed Class Specific Mean Autoencoder (2-layer) achieves a classification accuracy of \textbf{92.09\%}, which is at least 2.5\% better than existing algorithms. This is followed by VGG-Face with an accuracy of 89.45\%, while Face++ (commercial off-the-shelf system) achieves a classification performance of 78.41\%. The improvement of 5.29\% in performance of the proposed model  as compared to a Stacked Denoising Autoencoder can be attributed to the additional class representative terms added to the autoencoder formulation. 

Table \ref{tab:mixed} also presents the $p$-values obtained upon performing the McNemar test to evaluate the statistical difference. Since all the values are below 0.05, we can claim with a confidence level of 95\% that the performance of the proposed model is statistically different from all other existing models. In order to understand the effect of number of layers, experiments are also performed using a single layer Class Specific Mean Autoencoder. For a single layer, the proposed model yields an accuracy of 91.58\%, which continues to show an improvement of at least 2\%, compared to other models with the same architecture.

\begin{table*} [h]
\begin{center}
\small
\caption{Classification Accuracy (\%) on Multi-Ethnicity dataset. $p$-Value corresponds to the values obtained after performing McNemar test to compare the classification performance of an existing architecture with the proposed Class Specific Mean Autoencoder. The proposed model presents improved classification performance, while being statistically different from all other models at a confidence level of 95\%.}
\vspace{5pt}
\label{tab:mixed}
\begin{tabular}{ |m{7.5cm} | M{2.2cm}| M{2.2cm}|M{4cm}|} 
\hline
\textbf{Method} & \textbf{Accuracy (\%)} & \textbf{$p$-Value} & \textbf{Statistical Significance}\\ 
\hline
\hline
SDAE \citep{sdae} & 86.80 & 0.003 & Significant \\
\hline
DBM \citep{rbm} & 65.16 & $<$ 0.001 & Significant \\
\hline
DRBM \citep{drbm} & 87.03 & 0.001 & Significant \\ 
\hline
VGG-Face \citep{vgg} & 89.45 & 0.004 & Significant \\ 
\hline
COTS: Face++ \citep{faceplusplus} & 78.41 & $<$ 0.001 & Significant \\
\hline
\textbf{Proposed Class Specific Mean Autoencoder}  & \textbf{92.09} & - & - \\ 
\hline
\end{tabular}
\end{center}
\vspace{-15pt}
\end{table*}

\begin{table*}
\begin{center}
\small
\caption{Classification Accuracy (\%) on MORPH Album-II dataset. $p$-Value corresponds to the values obtained after performing McNemar test to compare the classification performance of an existing architecture with the proposed Class Specific Mean Autoencoder. The proposed model presents improved classification performance, while being statistically different from all other models at a confidence level of 95\%.}
\label{tab:morph}
\vspace{5pt}
\begin{tabular}{ |m{7.5cm} | M{2.2cm}| M{2.2cm}|M{4cm}|} 
\hline
\textbf{Method} & \textbf{Accuracy (\%)} & \textbf{$p$-Value} & \textbf{Statistical Significance}\\ \hline
\hline
SDAE \citep{sdae} & 66.25 & 0.005 & Significant \\
\hline
DBM \citep{rbm} & 65.30 & $<$ 0.001 & Significant \\
\hline
DRBM \citep{drbm} & 65.72 & $<$ 0.001 & Significant \\ 
\hline
VGG-Face \citep{vgg} & 70.44 & 0.010 & Significant \\ 
\hline
COTS: Face++ \citep{faceplusplus} & 57.23 & $<$ 0.001 & Significant \\
\hline
\textbf{Proposed Class Specific Mean Autoencoder} & \textbf{73.13} & - & - \\ 
\hline
\end{tabular}
\end{center}
\vspace{-15pt}
\end{table*}

\begin{figure}
\centering
\includegraphics[width= 3.3in]{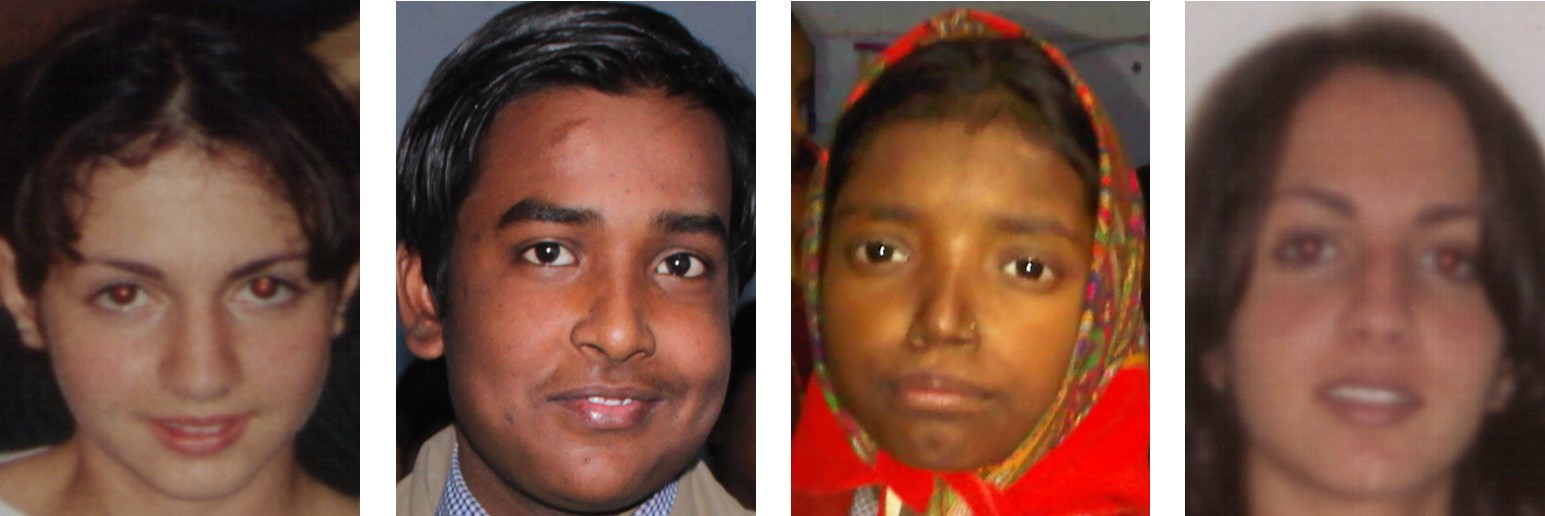} 
\caption{Sample images from the Multi-Ethnicity dataset, incorrectly classified by all algorithms. At the time of capture, all individuals were below the age of 18. It can be seen that while the actual age of the samples was below the age of majority, it is easy to mistake minors of 16-17 years as adults. External accessories such as scarves may also introduce mis-classification, resulting in unauthorized access control. }
\label{fig:kidsError}
\vspace{-10pt}
\end{figure}

\begin{figure*}
\centering
\subfloat[Multi-Ethnicity Dataset]{\includegraphics[width= 3.4in]{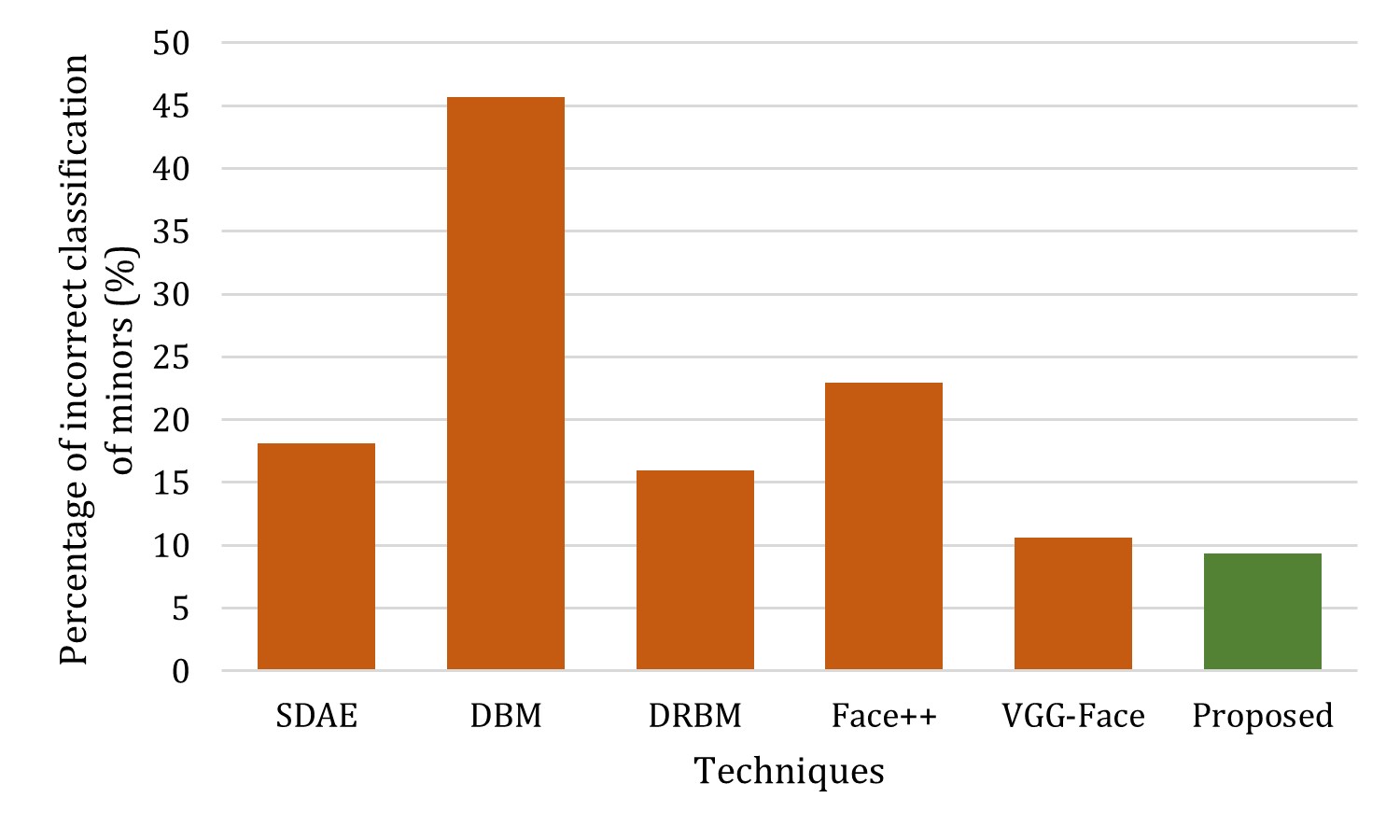}} 
\subfloat[MORPH Album-II Dataset]{\includegraphics[width= 3.4in]{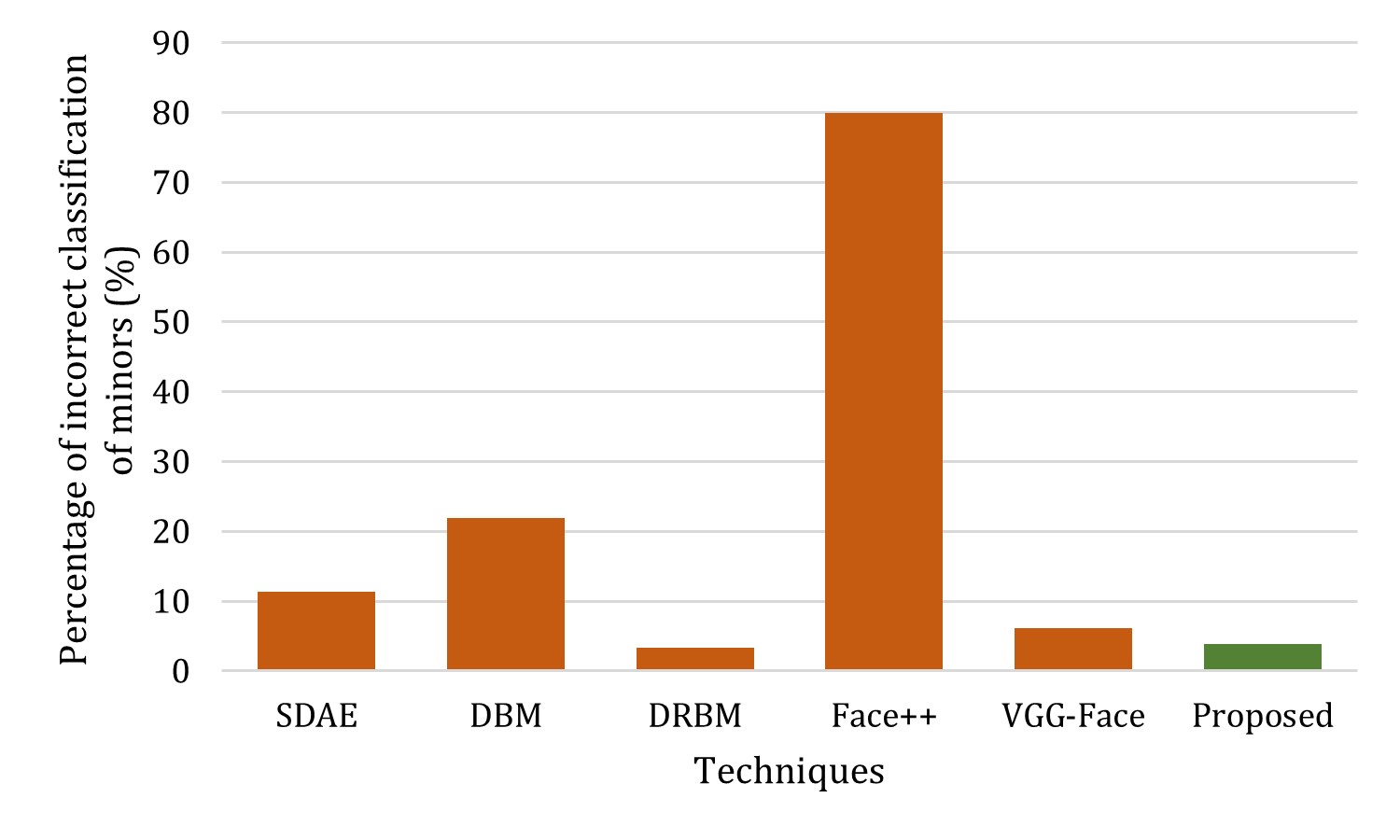}}  
\vspace{-5pt}
\caption{Percentage of minors incorrectly classified as adults for both the datasets. A lower percentage would ensure fewer instances of unauthorized access.}
\label{fig:bar}
\vspace{-10pt}
\end{figure*}

To analyze the class-specific classification accuracies, Table \ref{tab:conf} presents the confusion matrix for the proposed Class Specific Mean Autoencoder on the Multi-Ethnicity dataset. The results indicate that the performance of the trained model is not biased towards any particular class by achieving a classification accuracy of 93.52\% and 90.65\% on the two classes of adults and minors, respectively. This is essential to ensure that while unauthorized access is not provided to minors, rightful adults are not restricted from it either. In order to cater to the application of age-specific authorized access control, it is essential to ensure that the percentage of people below the age of majority, i.e. minor, obtaining unauthorized access should be minimal. To analyze the performance of all architectures for such an application, Fig. \ref{fig:bar} presents bar graphs summarizing the percentage of minors misclassified as adults. It can be seen that the proposed model achieves a misclassification percentage of 9.35\%, as opposed to 22.97\% by Face++. Fig. \ref{fig:kidsError} presents some sample images from the Multi-Ethnicity dataset misclassified as adults by \textit{all} the algorithms. It can be observed from the sample images that these images were captured either near the age of majority of 16-17 years or have artifacts such as headbands/scarves, which further make the task of adulthood classification challenging. Certain samples of kids below the age of one year were also mis-classified, possibly due to the undeveloped features of newborns.   

\begin{figure}
\centering
\includegraphics[width= 3.3in]{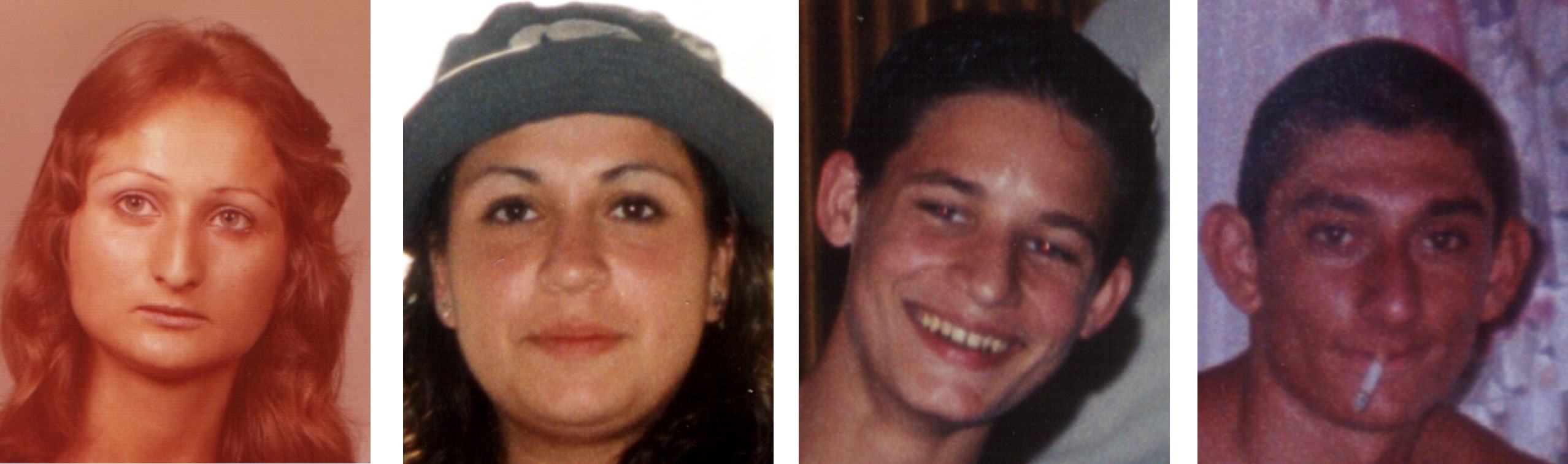} 
\caption{Sample images from the Multi-Ethnicity dataset that are in the age bracket of 16-19 years and misclassified by the proposed Class Specific Mean Autoencoder. The first image belongs to an adult of age 19 years, while the remaining belong to entities below the age of majority.}
\label{fig:fgnet}
\vspace{-10pt}
\end{figure}

\begin{table} [h]
\begin{center}
\caption{Confusion matrix of Class Specific Mean Autoencoder on the Multi-Ethnicity database.}
\label{tab:conf}
\begin{tabular}{ | M{0.3cm} || M{1.6cm} | M{1.6cm}|M{1.6cm}|  } 
\hline
& \multicolumn{3}{c|}{\textbf{Predicted}} \\ 
\hline
\hline
\multirow{3}{*}{\rotatebox[origin=c]{90}{\textbf{Actual}}}&  & Adult & Not Adult \\
\cline{2-4}
& Adult & 93.52\% & 6.48\% \\
\cline{2-4}
& Not Adult & 9.35\% & 90.65\% \\
\hline
\end{tabular}
\end{center}
\vspace{-15pt}
\end{table}

The major challenges associated with the problem of adult classification lie in the age bracket of 16 to 19 years (16-17: minors, and 18-19: adults). On the Multi-Ethnicity dataset, the proposed algorithm achieves a classification accuracy of 64.58\% on the above mentioned age range. VGG-Face, which performs the second best, reports an accuracy of 58.33\%, which is at least 6\% lower than the proposed algorithm. Fig. \ref{fig:fgnet} displays sample images from the specified age range and are misclassified by the proposed algorithm. The images demonstrate the challenging nature of human aging which are dependent on intrinsic and extrinsic person-specific factors, such as health, environment, and climate. 

\subsection{Results on MORPH Album-II Dataset}
The classification accuracies obtained by the proposed model and other existing architectures are tabulated in Table \ref{tab:morph}, and Fig. \ref{fig:roc} presents the Receiver Operating Characteristic (ROC) curves obtained for the experiments. It is observed that the proposed architecture achieves a classification accuracy of \textbf{73.13\%}, which is at least  2.5\% better than existing approaches, while Face++ (COTS) achieves an accuracy of 57.23\%. Table \ref{tab:morph} also presents the $p$-values obtained upon performing the McNemar statistical test on the proposed Class Specific Mean Autoencoder and other existing models. While the second best performance is achieved by VGG-Face features (70.44\%), it is important to note that the improvement in accuracy achieved by the proposed model is statistically significant for a confidence level of 95\%. Upon analyzing the gender-specific adulthood prediction results, it can be observed that the classification accuracy on female sample images is 62.89\%, whereas the accuracy on male sample images is 75.09\%. It is further observed that for females, the misclassification of adults as minors is much higher, as compared to males, thereby resulting in an overall lower classification performance. 

\begin{figure} [h]
\centering
\subfloat[Minors]{\includegraphics[height= 1in]{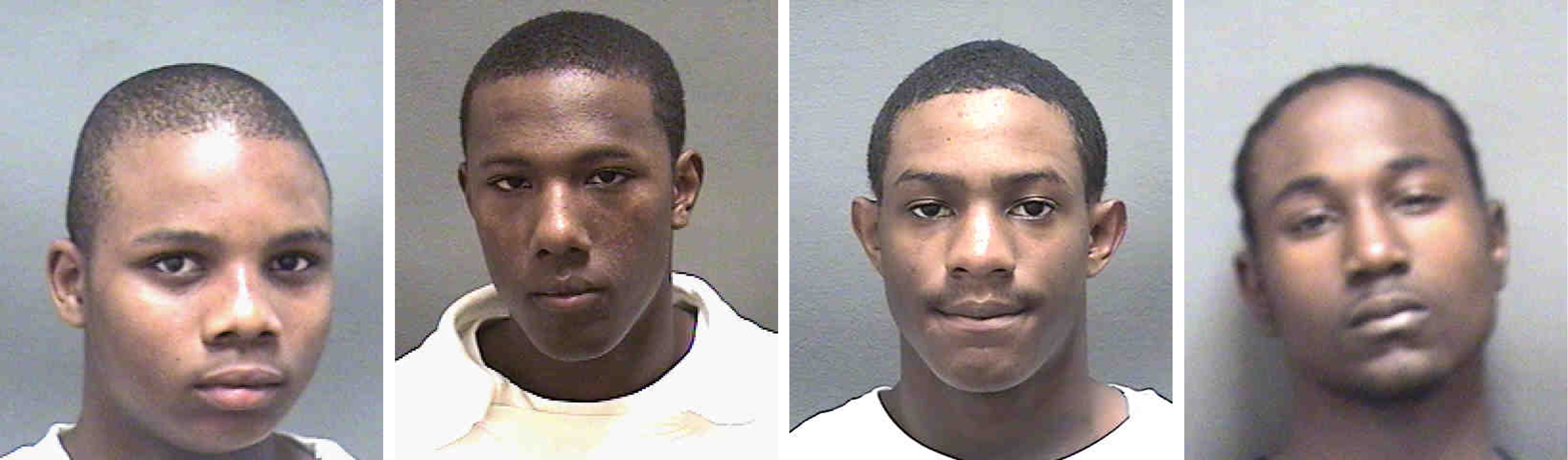}}  \\
\vspace{-5pt}
\subfloat[Adults]{\includegraphics[height= 1in]{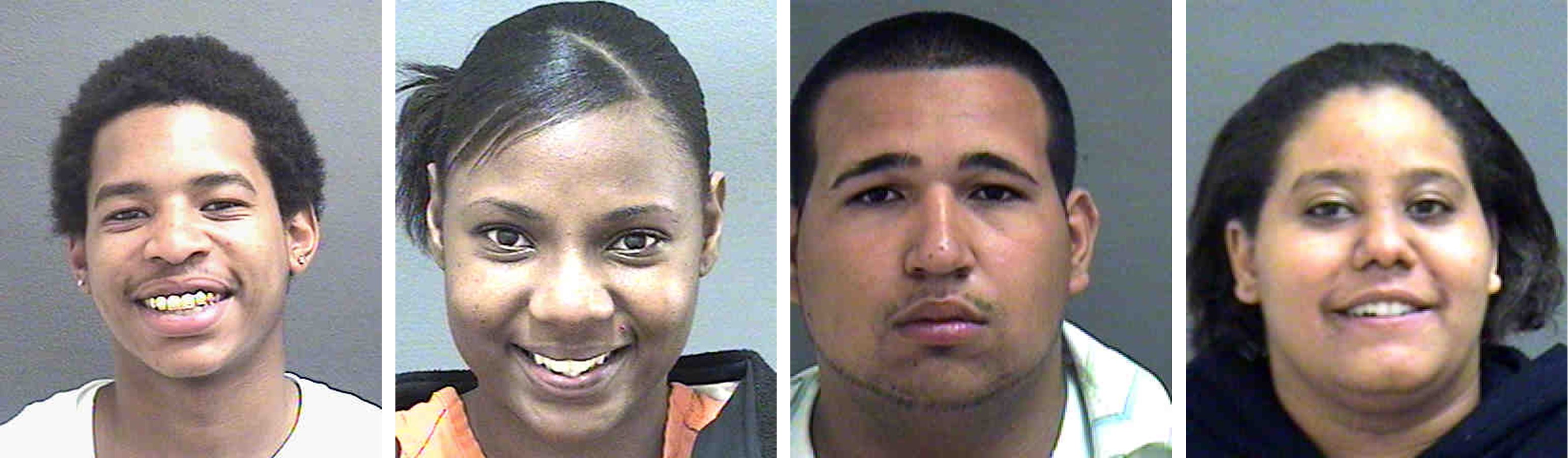}} 
\vspace{-5pt}
\caption{Sample images from the MORPH Album-II dataset correctly classified by the proposed algorithm, and not by other existing algorithms. (a) shows images of individuals having age 16 (first two samples) or 17, whereas (b) depicts just turned adults of 18 (first two) or 19 years of age.}
\label{fig:propCorrect}
\vspace{-10pt}
\end{figure}

\begin{table*} 
\begin{center}
\small
\caption{Classification accuracy (\%) on perturbed face images for Multi-Ethnicity and MORPH Album-II datasets.}
\label{tab:corrupted}
\vspace{5pt}
\begin{tabular}{|m{6cm} | M{2cm}| M{2cm}||M{2cm}|M{2cm}|} 
\hline
\multirow{2}{*}{\textbf{Perturbation}}& \multicolumn{2}{c||}{\textbf{Multi-Ethnicity Dataset}} & \multicolumn{2}{c|}{\textbf{MORPH Album-II Dataset}} \\
\cline{2-5}
 & \textbf{Proposed} & \textbf{VGG-Face} & \textbf{Proposed}& \textbf{VGG-Face} \\ \hline
\hline
No Perturbation (Original) & 92.09 & 89.45 & 73.13 & 70.44 \\
\hline
Gaussian Blur (Sigma = 3) & 89.54 & 61.34 & 72.40& 50.00  \\
\hline
Gaussian Noise (Mean = 0, Std. dev. = 0.01) & 87.95 & 50.59 & 72.09 & 50.03 \\ 
\hline
Gaussian Noise(Mean = 0, Std. dev. = 0.001) & 91.66 & 60.57 & 72.98 & 62.08 \\ 
\hline
Holes (10 holes of $3\times3$) & 87.35 & 64.67 & 72.55 & 56.70 \\
\hline
\end{tabular}
\end{center}
\end{table*}

From Fig. \ref{fig:bar}, it can be observed that the proposed model achieves a \textit{minor} misclassification percentage of only 3.9\%, as opposed to nearly 80\% by Face++ (COTS) on the MORPH Album-II dataset. The high misclassification rate of minors by Face++ reinstates the requirement for robust algorithms with the ability to process and analyze minor face images as well. It is important to note that the age of face images in MORPH Album-II dataset varies from 16 to 77 years. Thus, resulting in a further challenging dataset having multiple subjects \textit{just below} the age of majority. The higher misclassification rate achieved can thus also be attributed to this challenging age range. It is also interesting to observe from Fig. \ref{fig:bar} that while DRBM achieves a lower misclassification of minors as adults (i.e. 3.30\%), the overall classification accuracy of DRBM based approach is less than the proposed approach (Table \ref{tab:morph}). This further motivates the use of the proposed algorithm for ensuring rightful access to adults, while restricting minors. 

Fig. \ref{fig:propCorrect} presents sample images of the age group of 16-19 (16-17: minors, 18-19: adults) years from the MORPH Album-II dataset which are correctly identified by the proposed algorithm and not by any other algorithm. Upon analyzing the mean images of both the classes, we observe a significant visual difference in the jaw area of minors and adults. As can be seen from Fig. \ref{fig:propCorrect} as well, minors appear to have a tighter jaw line which is often not observed with adults. We believe that this variation has been encoded well by the proposed model among other features, resulting in superior performance. 

\subsection{Performance on Perturbed Face Images}
It has often been observed in literature that the performance of deep models deteriorates in the presence of perturbations \citep{adversary}. The proposed model has also been evaluated on perturbed face images in order to understand its vulnerabilities. This is performed by incorporating perturbations in the form of Gaussian blur, Gaussian noise, and holes in the original face images. Experiments are performed on the Multi-Ethnicity and the MORPH Album-II datasets with the protocols discussed earlier. The models are trained on unperturbed (original) images but the test images are perturbed. In this evaluation, no separate training is performed for the perturbed face images. Table \ref{tab:corrupted} presents the classification accuracies obtained from the proposed Class Specific Mean Autoencoder, and the second best performing model, VGG-Face. It can be observed that with perturbed test images, the accuracy of the proposed model reduces by less than 5\% and 1.04\% for Multi-Ethnicity and MORPH Album-II datasets, respectively. On the other hand, VGG-Face demonstrates a drop of at least 24\% and 8\% on the two datasets, respectively. This experiment demonstrates the utility of the proposed model for performing classification under different kinds of perturbations.


\section{Conclusion}
Faces are often seen as a viable non-invasive modality for predicting the age of an individual. However, due to the large intra-class variations, predicting adulthood from face images is an arduous task. The key contribution of this research is developing a novel formulation for \textit{Class Specific Mean Autoencoder} and utilize it for adulthood classification. The proposed formulation aims to learn supervised feature vectors that maximize the intra-class similarity.  Experimental results and comparison with existing approaches on two large databases: the proposed Multi-Ethnicity dataset and MORPH Album-II dataset showcase the effectiveness of the proposed algorithm. In future, we plan to extend the proposed formulation to incorporate multiclass-multilabel information in feature learning. 

\bibliographystyle{model2-names}
\bibliography{bibFile}


\end{document}